\documentclass[twocolumn,10pt]{article}
\usepackage[utf8]{inputenc}

\usepackage[letterpaper,top=2.75cm,bottom=2.75cm,left=1.8cm,right=1.8cm,marginparwidth=1.75cm]{geometry}

\usepackage{cite}
\usepackage{xcolor}
\usepackage{amsmath,amssymb,amsfonts}
\usepackage{algorithm}
\usepackage[noend]{algorithmic}

\usepackage{graphicx}

\usepackage{pgffor}
\usepackage{import}
\usepackage{subcaption}
\usepackage{multirow}
\usepackage{bbding}
\usepackage{authblk}

\usepackage{caption}
\usepackage{siunitx}
\usepackage[strings]{underscore}
\usepackage[hidelinks]{hyperref}
\usepackage{microtype}

\def\CC{{C\nolinebreak[4]\hspace{-.05em}\raisebox{.4ex}{\tiny\bf ++}}}

\title{Road to scalability for efficient graph search on massively parallel neuromorphic hardware\\
}

\author[1,2]{Oskar von Seeler}
\author[3]{Elena Offenberg}
\author[3]{Carlo Michaelis}
\author[3]{Tomas Kulvicius}
\author[1,2,3,4]{Jannik Luboeinski}
\author[1,2,3,5,*]{Andrew B. Lehr}
\author[1,2,3,*,\Envelope]{Christian Tetzlaff}

\affil[1]{\small Department of Neuro- and Sensory Physiology,
University Medical Center G{\"o}ttingen, Germany}
\affil[2]{\small Campus Institute Data Science,
University of G{\"o}ttingen, Germany}
\affil[3]{\small Third Institute of Physics -- Biophysics, University of Göttingen, Germany}
\affil[4]{\small J\"{u}lich Supercomputing Centre, Forschungszentrum J\"{u}lich, Germany}
\affil[5]{\small Circulant Labs,
Bensheim, Germany}
\affil[*]{co-senior authors}
\affil[\Envelope]{Correspondence: christian.tetzlaff@med.uni-goettingen.de}

\date{}

\begin{document}

\maketitle

\null

\begin{abstract}

Efficient computation of shortest paths in weighted graphs is a fundamental problem with many applications. Neuromorphic hardware platforms promise massively parallel, efficient computation, changing parallelism tradeoffs.
In this work, we introduce NEURO-MAPP (Neuromorphic-based Min-Add Parallel Propagation), a distributed shortest path algorithm designed to use the local computation and network communication available in neuromorphic systems. We provide an optimized implementation of the algorithm on the SpiNNaker 2 platform and evaluate its performance on a selection of synthetic and real-world graphs. These results are compared to Dijkstra's algorithm on a modern CPU. We find that the NEURO-MAPP implementation scales favorably in terms of runtime for many graph types while consuming less energy per shortest-path query than the CPU implementation in almost all cases. 
These findings highlight the potential of neuromorphic hardware featuring sparse, spike-based communication as a scalable and energy-efficient platform for computation in graph search and related tasks.

\end{abstract}

\section{Introduction}\label{sec.introduction}

The optimal traversal of a weighted graph is a fundamental problem in mathematics and computer science with applications in multiple disciplines. It arises whenever a system can be described as a set of states or actions that are connected by possible transitions. 
The computation of shortest paths is critical in many everyday settings, such as route planning for map-based navigation applications, efficient signal routing in computer networks and power grids, robotic action planning, as well as the analysis of large, high-dimensional data sets through nonlinear dimensionality reduction techniques like Isomap~\cite{tenenbaum_global_2000}.

A variety of algorithms have been developed that solve the shortest-path problem on conventional hardware in a sequential manner, for example Dijkstra’s Algorithm \cite{dijkstra_note_1959}, the Floyd-Warshall Algorithm \cite{floyd_algorithm_1962, warshall_theorem_1962}, and the Bellman-Ford Algorithm \cite{ford_network_1956,bellman_routing_1958} (also cf. \cite{kumawat_extensive_2021,magzhan_review_2013} for reviews). Building on earlier theoretical works \cite{kumar1991scalability,meyer2003delta} and the availability of GPU-based computing, algorithms have been designed utilizing parallel computing architectures to accelerate graph search tasks~\cite{wang_gunrock_2016,wang_fast_2021,davidson_work-efficient_2014,busato_efficient_2016}, in addition to novel solutions like genetic algorithms~\cite{karas_genetic_2011,zhang_hybrid_2018} or artificial neural networks~\cite{kulvicius_finding_2022, kulvicius_combining_2024}.

The substantial computational demands associated with large-scale graph search problems, arising from both graph size and multi-user production environments, underscore the need for solutions that optimize runtime, energy efficiency, and scalability.
Recent brain-inspired neuromorphic hardware provides particular capabilities for highly efficient execution of spiking neural networks \cite{davies_advancing_2021,yan2021, pehle2022brainscales,hoppner_spinnaker_2022}.
On these neuromorphic hardware systems, computation and memory are typically co-localized and event-based communication is used to enable the energy-efficient and scalable implementation of networks of neurons and synapses.

\begin{figure*}
    \centering\includegraphics[width=\textwidth]{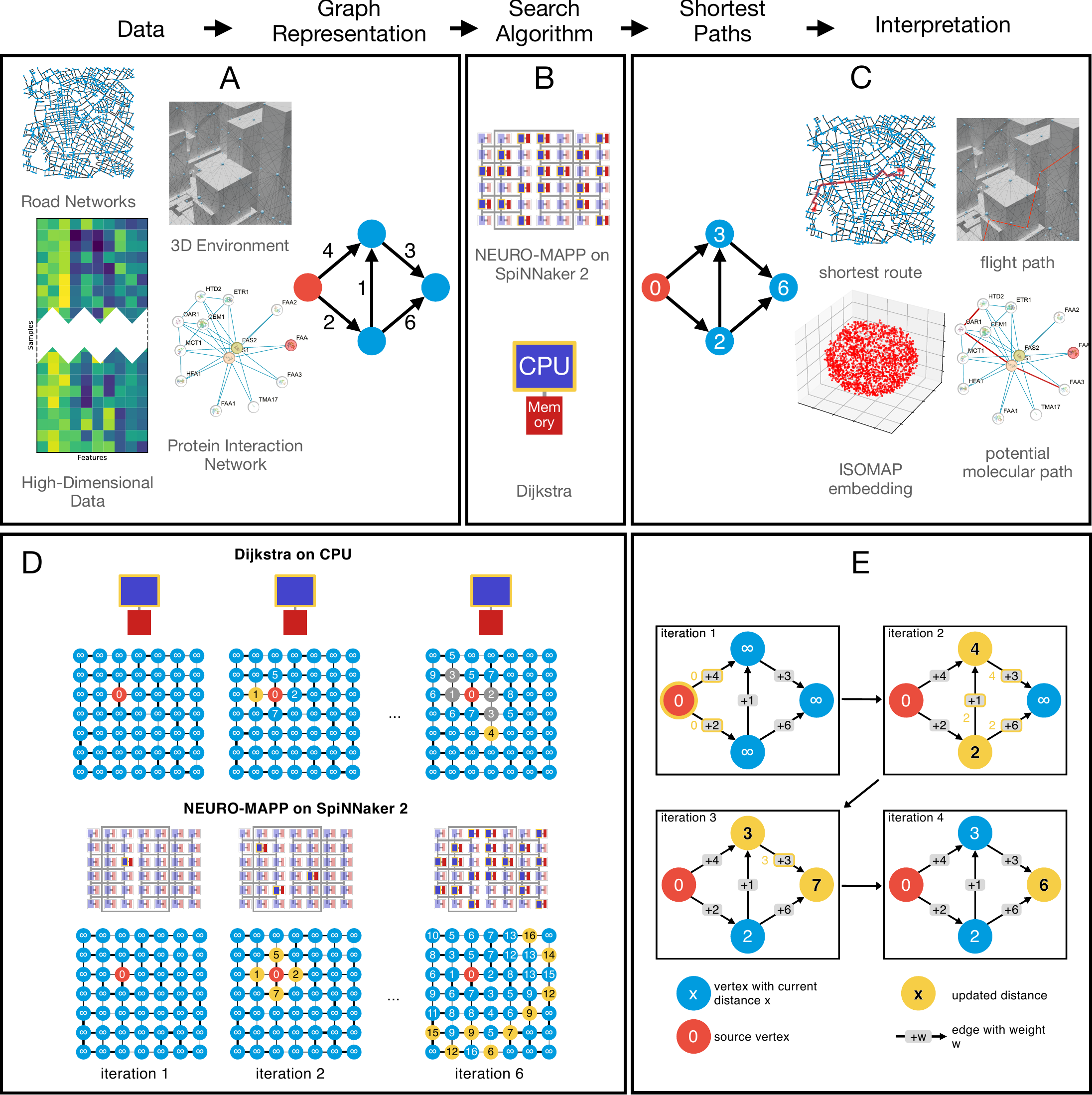}
        \caption{Overview of this work. \textbf{A} Multiple real-world problems can be formulated as shortest path queries on weighted graphs. \textbf{B} In this work, two options for finding shortest paths in the graphs are used and compared: the NEURO-MAPP algorithm on SpiNNaker 2 and Dijkstra's algorithm on a CPU. \textbf{C} Solutions of the shortest path problem resulting from either algorithm. \textbf{D} Visualization of iteration in Dijkstra's algorithm and NEURO-MAPP exemplified on a 2D grid graph. Edge weights are indicated by thickness. Source vertices are colored red, currently visited vertices yellow. For Dijkstra's algorithm, previously visited vertices are gray. \textbf{E} Schematic of the algorithmic sequence of NEURO-MAPP on a small example. Currently visited vertices and the weights of their outgoing edges are highlighted in yellow.}
        \label{fig:am_schematic}
\end{figure*}

Although theoretical frameworks based on spiking neural networks suggest a promising route toward efficient implementations of graph search algorithms on neuromorphic hardware~\cite{aibara_cmos_1991,ponulak_rapid_2013,davies_advancing_2021,aimone_provable_2021}, established graph search approaches such as Dijkstra’s Algorithm and the delta-stepping algorithm remain computationally demanding (cf.~\cite{davies_advancing_2021,hoppner_spinnaker_2022}). This limitation arises primarily from their reliance on global coordination mechanisms, which can be costly in distributed or event-driven architectures such as neuromorphic hardware.

To address this challenge, we propose an algorithm that leverages the inherently inexpensive local computations characteristic of neuromorphic hardware systems while minimizing the need for global synchronization. Conceptually, our approach is related to the class of Parallel Multiple-Queue algorithms~\cite{bertsekas_parallel_1996}, adapting their principles to better align with the architectural constraints and strengths of neuromorphic hardware.

Specifically, we introduce \textit{NEURO-MAPP (Neuromorphic-based Min-Add Parallel Propagation)},
an iterative algorithm that embeds a weighted graph into a corresponding computational graph for shortest path finding on neuromorphic hardware. 
In this representation, edges propagate distance updates by adding their associated weights, while vertices perform local minimization over incoming signals to select the smallest candidate distance. Through this distributed min-add operation, shortest paths emerge as a consequence of purely local, parallel update dynamics, without requiring global coordination.

We demonstrate that implementing this algorithm on massively parallel neuromorphic hardware -- specifically, a SpiNNaker 2 system -- enables graph search to take place at all vertices/edges in parallel (see Figure~\ref{fig:am_schematic}).
This enables highly efficient computation of shortest paths as well as fully parallel computation of distances from all locations to nearest points of interest.
To achieve efficient implementation of NEURO-MAPP on SpiNNaker 2, our work further includes new approaches for graph partitioning and on-chip scheduling, tailored to the hardware's specifications. 
Generally, NEURO-MAPP could be implemented on any system with sufficiently many independent parallel computational units, provided that each unit has sufficient local memory to store all data required for the assigned graph section, and there is a communication network that allows to send short messages with low latency between these units.

\section{Results}\label{sec.results}

\begin{figure*}
    \centering
    \includegraphics[width=0.9\linewidth]{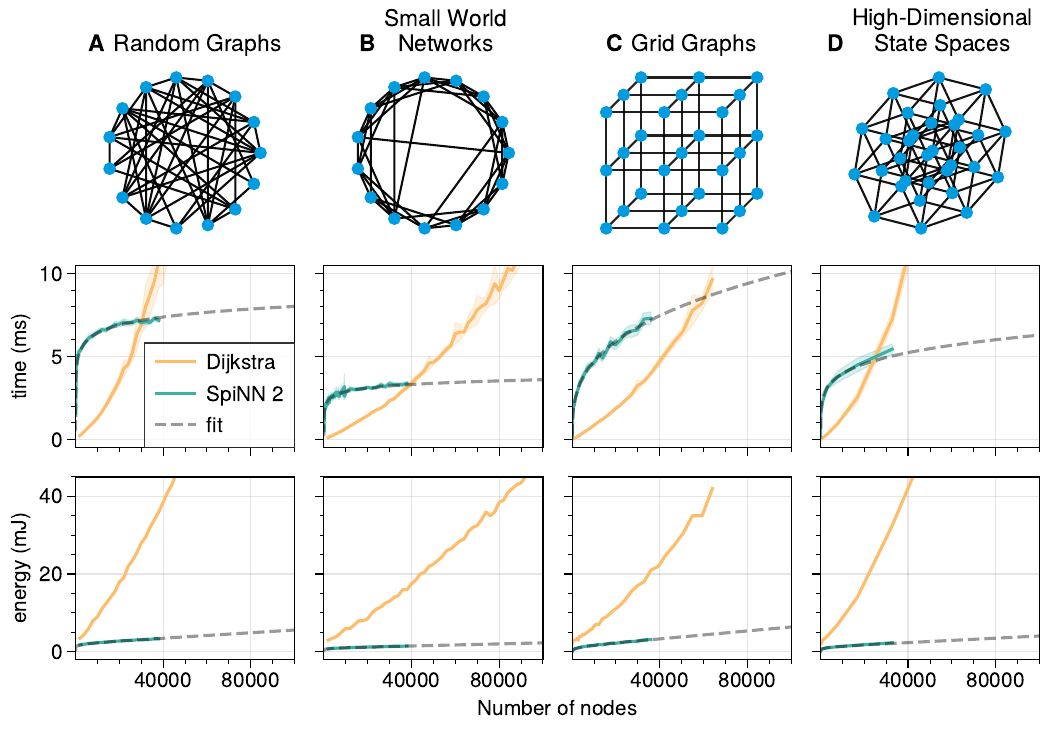}
    \caption{Runtime performance and energy consumption measurements for four different, generic graph types. For all graphs a fixed set of parameters is chosen and $|V|$ is varied systematically. First row shows a visual representation of the graph type. Second row shows runtime for NEURO-MAPP (green) and Dijkstra's algorithm (yellow). For both implementations, 20 trials were executed for each set of parameters. The colored lines show the mean of these executions with the 95\% confidence interval being displayed as a transparent band. Beyond the capacity of a single chip, an estimate for the runtime on chip is extrapolated from the available measurements (see main text for more details). The estimate is shown with gray dashes, including the fit of the data. Third row shows the results of energy measurement for NEURO-MAPP (green) and Dijkstra's algorithm (yellow). \textbf{A} Random graphs with fixed out-degree of 12. \textbf{B} Small-world graph following the Watts-Strogatz model with $k=4$ and $p=0.1$. \textbf{C} A 3D grid graph with maximal $33^3 = \num{35937}$ vertices and \num{209088} edges. \textbf{D} 5D grid graph.}
    \label{fig:F2}
\end{figure*}

Here we set out to efficiently compute shortest distances between vertices in weighted, directed graphs of various underlying statistics using massively parallel neuromorphic hardware.
In brief, NEURO-MAPP converts a weighted graph into a computational graph, whose edges add their weight to passing distance updates and whose vertices compute the minimum of incoming distance updates and their internal state (Fig.~\ref{fig:am_schematic}D,E). Further, NEURO-MAPP splits the computational graph into subsets of vertices that are mapped onto the cores of a SpiNNaker 2 chip (cf. Fig.~\ref{fig:am_schematic}D). 
A detailed description of NEURO-MAPP and its implementation on SpiNNaker 2 is provided in the Methods (Section~\ref{subsec.algorithm}).

A single SpiNNaker 2 chip provides 152 ARM cores, each forming a so-called processing element together with its own fast SRAM memory~\cite{hoppner_spinnaker_2022}. In addition, one DRAM module with \qty{2}{\giga\byte} of memory is attached to the chip. 
The cores are connected to each other through a communication network designed for sending many small messages, which we utilize and adapt for sending distance updates between vertices.
Here we measure the performance characteristics of a single chip with maximal graph sizes of around \num{38000} vertices (see section~\ref{sec:met:impl:part} for details). As in the future, multiple SpiNNaker 2 chips may be connected to form larger networks, we further consider according extrapolations of our results.

To achieve the performance results presented in the following, we had to make some key optimizations, including vertex-to-core assignment and idle time reduction.

\textit{Vertex-to-core assignment:} We have found that the distribution of vertices across cores has a significant impact on the performance. To optimally utilize the parallel architecture of SpiNNaker 2, it is essential to ensure that vertices are assigned to cores in a manner that evenly distributes computational effort.
In contrast to spiking neural networks previously deployed on SpiNNaker 2, some of the graphs being considered here are comparatively sparse and have strongly skewed distributions of vertex degrees, such as protein interaction networks, which exhibit ``hubs'' with high connectivity (Supplementary Figs.~8,9). To consider these graph properties, we changed the computational format in which the graph is stored in the SpiNNaker 2 codebase. This change slightly increased the memory required per edge, but substantially reduced the processing time required for each processed update (cf. Section~\ref{methods.impl.mp}).
Therefore, for the 3D navigation task and the protein interaction networks we used this vertex-degree-aware optimized partitioning, while for the abstract graphs shown in Fig.~\ref{fig:F2} as well as for the road navigation and Isomap tasks we used randomized assignment of vertices to cores.

\textit{Idle time reduction:} The algorithm is executed iteratively such that in each iteration incoming distance updates are processed and new updates are triggered in the eligible vertices. The runtime for each iteration can vary for each processing element as well as over time, reflecting differences in local workload.
To reduce idle time between iterations, a lightweight global coordination mechanism is required. We implemented this as follows: Once a core completes its workload for the current iteration, it sends a signal to a global coordinator. After receiving completion signals from all cores, the coordinator starts the next iteration by sending a message to all processing elements (Supplementary Fig.~18, for details cf. Section~\ref{sec:dynamic}). 

\subsection*{Superior scaling behavior of NEURO-MAPP across various graph types}

In the present study, we compare NEURO-MAPP, as implemented on a single SpiNNaker 2 chip that comprises many low-power cores, to Dijkstra's algorithm executed on a single, conventional CPU core. As both runtime performance and the energy efficiency are expected to depend on graph size (measured by the number of vertices), edge density, and topology of the graphs, we systematically evaluate and compare both implementations with respect to these characteristics.

The iterative nature of NEURO-MAPP implies that it requires as many iterations as the maximum number of edges that needs to be traversed for any shortest path originating from the source vertex. Although the computational cost per iteration depends on graph-specific properties, the total number of iterations exerts the dominant influence on overall runtime. Therefore, we assume that the diameter of the corresponding unweighted graph, which is defined as the maximum number of edges required to connect any two vertices, provides an approximation of the number of iterations needed. Empirically, for the graph classes examined in this work, the number of iterations scales approximately linearly with the unweighted graph diameter (see Supplementary Figs.~1-6).
Consequently, favorable performance is observed for graph families whose (unweighted) diameter is low such as highly dense graphs, but also certain sparse graphs such as small-world networks, multidimensional grids, or randomly connected graphs possess relatively low diameters.

To investigate this scaling behavior and to benchmark our approach against Dijkstra's algorithm, we evaluated both implementations on a diverse set of generic graph types (Figure~\ref{fig:F2}). A comprehensive description of graph generation procedures and used parameter settings is provided in the Methods (section~\ref{sec.methods.graphs}).

\textit{Random graphs.} We start by evaluating the algorithms' performances on random graphs. In this scenario, we expect a superior performance of NEURO-MAPP, as activity should propagate rapidly across all vertices, enabling convergence within a relatively small number of iterations. As shown in Figure~\ref{fig:F2}A, the Dijkstra implementation performs better for small graphs; however, as the number of vertices increases, the number of iterations needed for NEURO-MAPP grows more slowly such that for graphs with more than \num{30000} vertices and \num{360000} edges NEURO-MAPP shows superior performance. At the maximum graph size that can be implemented on a single SpiNNaker 2 chip (\num{38000} vertices), NEURO-MAPP is about $ 25 \%$ faster.
Additional experiments verify this finding across different parameters of the random graphs (Supplementary Fig.~6).
Based on our results from graphs that fit on a single chip, we further estimate that the runtime increases logarithmically with the graph size for random graphs. A logarithmic model $t \sim a \log(|V|+1)$ fits the measured data with $R^2 = 85.8\%$ (see Fig.~\ref{fig:F2}A and Supplementary Fig.~12), supporting the finding that NEURO-MAPP solves the shortest-path problem in random graphs substantially faster.

\textit{Small-world graphs}. Many real-world networks exhibit small-world characteristics, typically combining short average path lengths with strong local clustering. Representative examples include power grids, social networks and brain networks~\cite{amaral_classes_2000}.
We generated small-world graphs using the Watts-Strogatz model~\cite{watts_collective_1998}.
Figure~\ref{fig:F2}B shows that, similar to random graphs, the runtime of NEURO-MAPP grows logarithmically with increasing number of vertices ($R^2=89.6\%$, Supplementary Fig.~13), outperforming the Dijkstra implementation for graphs with more than \num{38000} vertices. This logarithmic dependency holds across variations of typical parameters of small-world graphs such as the edge rewiring probability $p$ (see Supplementary Fig.~3) and the fixed number of vertex neighbors $k$ (see Supplementary Fig.~1).
In general, decreasing $p$thereby reducing the number of random long-range edges results in slower operation of NEURO-MAPP, while slightly improving the performance of Dijkstra's algorithm (see Supplementary Fig.~3).
This slowdown of NEURO-MAPP arises from the increased number of iterations required to propagate activity throughout the entire graph, whereas the improvement in Dijkstra's algorithm likely stems from a smaller search frontier and therefore reduced heap sizes. In the limit case of $p=0$, where no long-range connections are present, the resulting graph forms a uniform ring structure. Such graphs exhibit a diameter that grows linearly with the number of vertices. Under these conditions, NEURO-MAPP does not exhibit sub-linear scaling in either iteration count or runtime, leading to performance advantages for Dijkstra's algorithm (Supplementary Fig.~4). Aside from such limit cases, also for small-world graphs NEURO-MAPP shows superior runtime performance with increasing graph size.

\textit{Multidimensional grid graphs.} Grid-structured graphs can represent physical space in two or three dimensions, and may incorporate time as an additional dimension. More generally, high-dimensional grid graphs can be used to represent the state spaces of high-dimensional physical systems, such as those associated with robotic joint configurations.
Figure~\ref{fig:F2}C shows results for three-dimensional grid graphs with randomly assigned weights. In these graphs the diameter, and therefore the number of iterations required for convergence, grows faster with graph size than in random or small-world graphs. Consequently, for all three-dimensional grid graphs that fit on a single chip, NEURO-MAPP remains slower than Dijkstra’s algorithm, although the difference diminishes for the largest graphs, where the runtime is less than twice as large.
For higher-dimensional grid graphs, however, a performance advantage emerges for NEURO-MAPP (see Supplementary Fig.~5). For example, a five-dimensional grid with $8^5 = \num{32768}$ vertices and \num{286720} edges can be searched about $20\%$ faster using NEURO-MAPP (see Figure~\ref{fig:F2}D).
Curve fits (Fig.~\ref{fig:F2}C,D for $d=3$ and $d=5$, Supplementary Fig.~11 for $d \in \{2,3,4,5\}$) with $R^2 \approx 90\%$ suggest that the runtime for the considered multidimensional grid graphs scales with $\mathcal O(\sqrt[d]{n})$. These results indicate that, for sufficiently large grid graphs, NEURO-MAPP is expected to outperform Dijkstra’s algorithm.

\subsubsection*{Energy Consumption}

In addition to the runtime, we evaluated the energy consumption of NEURO-MAPP on the SpiNNaker 2 chip. For this, power draw was measured independently of the runtime by repeatedly executing the graph search for 5 seconds for each input, allowing consistent measurements (see Methods for more details). Across all four graph types, the power draw grows approximately linearly with graph size. A linear model $P \sim a|V|+b$ fits the measurements with $R^2>99\%$ for each graph type (see Supplementary Fig.~16), likely caused by the increasing number of cores used during execution.
Energy consumption was calculated by multiplying the average power draw with the previously measured runtime of a single search. 
For all four graph types, NEURO-MAPP is more energy-efficient than the CPU-based Dijkstra algorithm, including for smaller graphs where Dijkstra's algorithm requires lower runtime (see Fig.~\ref{fig:F2} third row).
Because power consumption scales approximately linearly with graph size, there may nevertheless exist graph regimes where NEURO-MAPP becomes less energy-efficient than Dijkstra’s algorithm.
For random graphs, assuming a runtime scaling of $\mathcal O(\log n)$ and power scaling of $\mathcal O(n)$, the resulting energy consumption is expected to scale as $\mathcal O(n \log n)$, which is comparable to Dijkstra's algorithm. For $d$-dimensional grid graphs, the runtime is assumed to scale as $\mathcal O(n^{1/d})$, resulting in energy consumption scaling as $ \mathcal O(n \cdot n^{1/d})$, which exceeds in general $\mathcal O(n \log n)$ of Dijkstra's algorithm.
In practice, however, the relatively low maximum power draw for graphs that require the usage of all cores of the SpiNNaker 2 chip (less than \qty{500}{\milli\watt}, see Supplementary Fig.~16, compared to around \qty{4}{\watt} for the CPU) suggests that this regime would only become relevant for very large graphs.

\subsection*{Performance of NEURO-MAPP across various applications}

Our evaluation of NEURO-MAPP on generic graph types has demonstrated favorable scaling behavior, with lower runtime and energy consumption, compared to Dijkstra's algorithm. Building on these results, we next assess the applicability of NEURO-MAPP across several representative use cases. Specifically, we apply NEURO-MAPP to i) classical road-network navigation, ii) aerial path planning in a three-dimensional environment, iii) the analysis of protein interaction networks, and iv) its integration as a component within a larger algorithmic framework, by the example of the dimensionality reduction method ISOMAP.

\subsubsection*{Road network navigation}

\begin{figure}
    \centering
    \includegraphics[width=\linewidth]{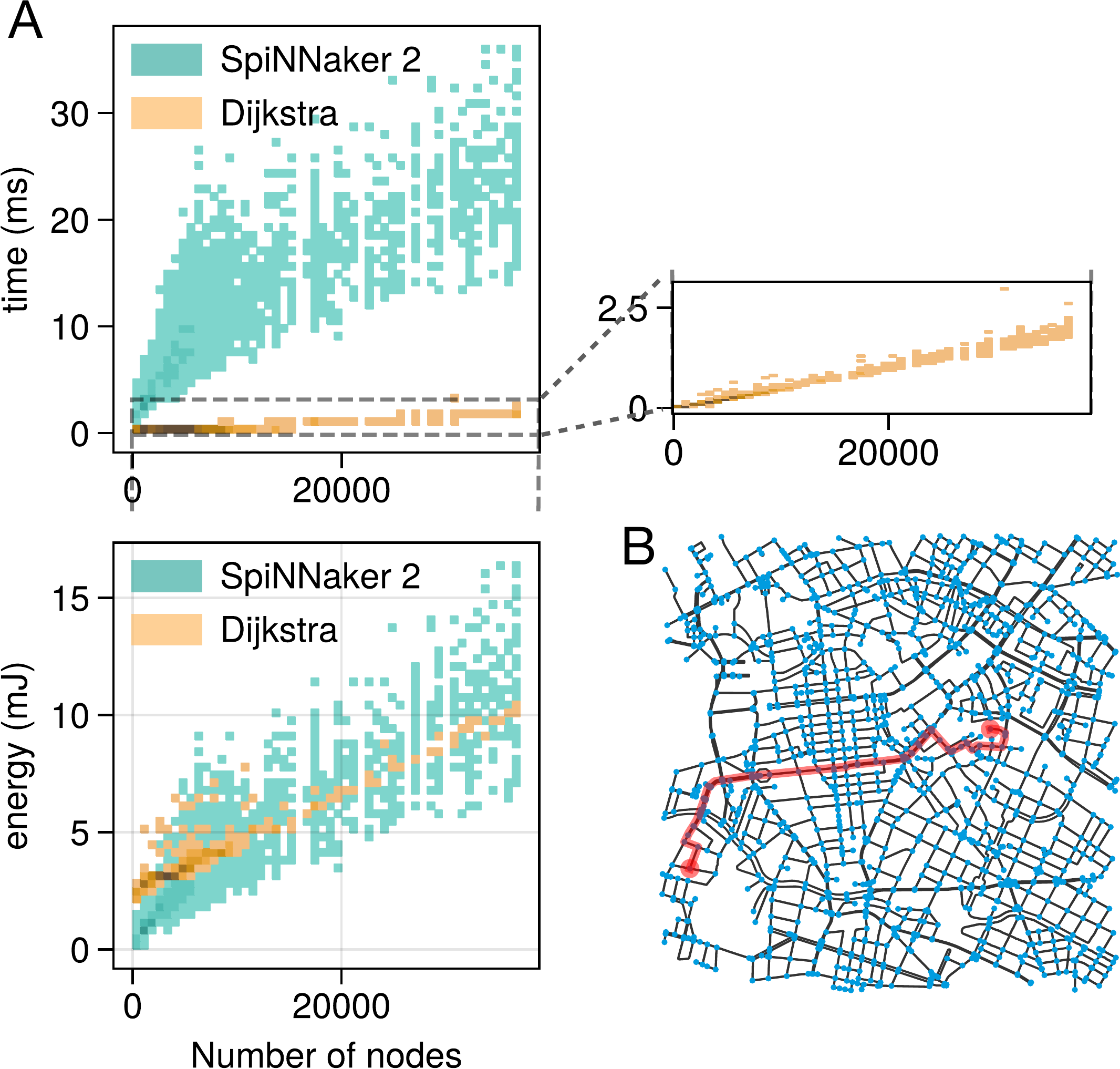}
    \caption{Navigation in road networks from Germany.
    \textbf{A} Runtime measurements for 486 different road networks taken from different areas in Germany for NEURO-MAPP (green) and Dijstra's algorithm (yellow). \textbf{B} Example of a road network. Blue dots are the vertices of the graph, which represent intersections in the road network. The edges (black) are road segments connecting intersections. One exemplary shortest path is drawn in red.}
    \label{fig:roads}
\end{figure}

A typical real-world application for evaluating the performance of SSSP algorithms is the search for optimal paths in road networks. For this, we again evaluated NEURO-MAPP on SpiNNaker 2 and compared it to Dikstra's algorithm on CPU, using exemplary road networks from Germany derived from OpenStreetMap (OSM). As shown in Fig.~\ref{fig:roads}A, for the network sizes considered, NEURO-MAPP is approximately one order of magnitude slower than Dijkstra's algorithm.

However, when comparing the scaling with increasing network sizes to the trends observed for the generic graph types (Fig.~\ref{fig:F2}), it is expected that NEURO-MAPP will surpass Dijkstra's algorithm for sufficiently large networks. Running NEURO-MAPP for such network sizes will require deployment on a multi-chip SpiNNaker 2 system.

\subsubsection*{Aerial navigation in an urban setting}

\begin{figure}
    \centering
    \includegraphics[width=\linewidth]{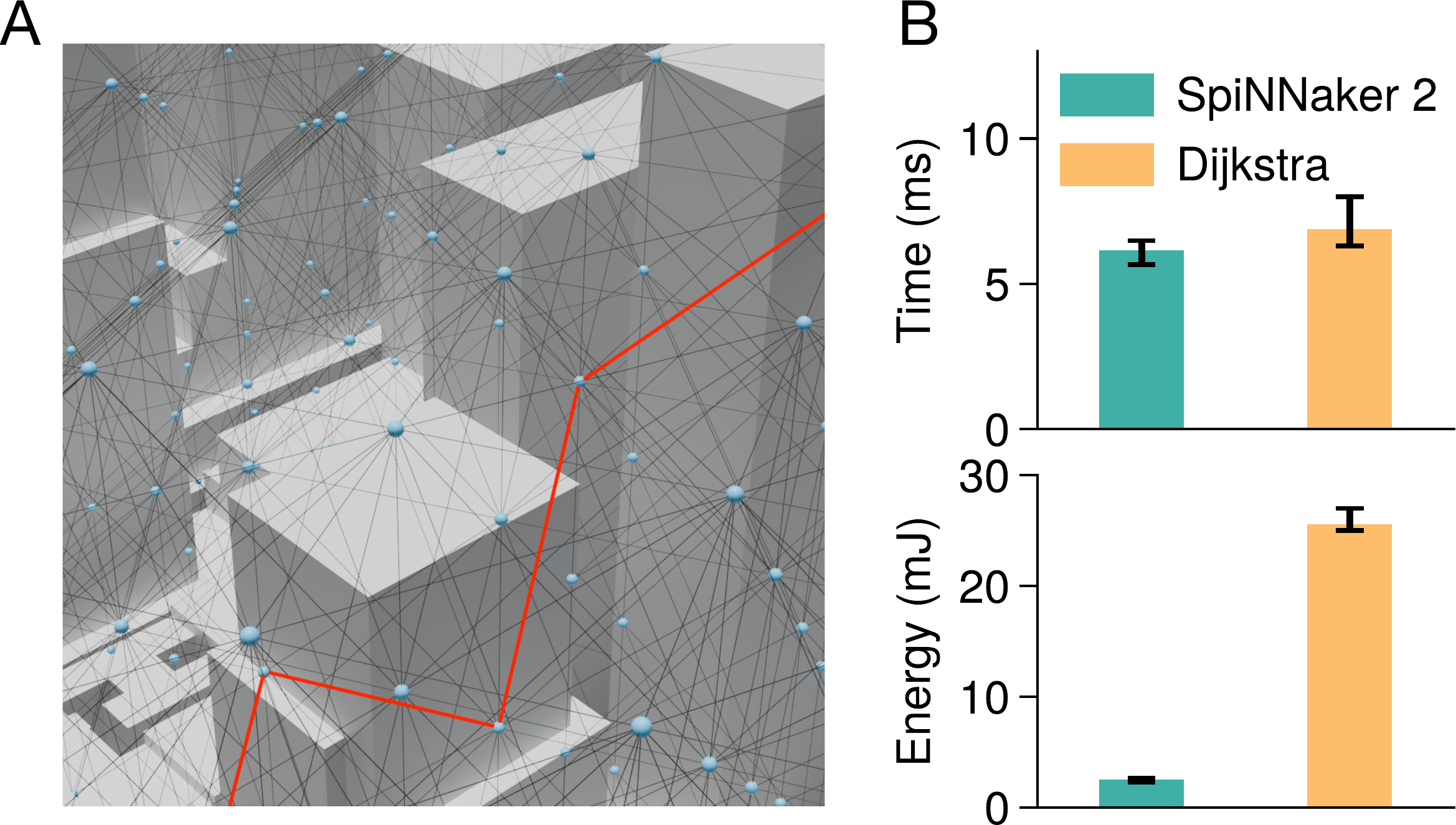}
    \caption{Shortest path finding for aerial navigation in a three-dimensional urban environment with $38392$ vertices and $639582$ edges. \textbf{A} Visualization of a small graph section with buildings and the related network. \textbf{B} Runtime and energy consumption for NEURO-MAPP and Dijkstra's algorithm for the full 3D graph. Plots show median and interquartile range over 20 trials.
    }
    \label{fig:3dgraphs}
\end{figure}

Similar to road network navigation, path-finding for aerial vehicles like drones in urban environments can be formulated as a SSSP problem on a network representation of the three-dimensional city space.
We constructed such a network by utilizing data from OpenStreetMap (OSM), which provides detailed geometric information on buildings, including footprints, approximate heights, and roof structures. This information enables a reliable reconstruction of the 3D urban environment.
More specifically, we generate a graph from the reconstructed 3D model by embedding a regular grid of points within the navigable airspace, where each point corresponds to a vertex. Edges are established between adjacent points, including diagonal neighbors, if the direct line between these two points is not obstructed by a building (see Methods section~\ref{sec.methods.graphs.real} for further details). 
We considered here as example an area of \qty{2.6}{\square\kilo\meter} in Manhattan and transformed it into a network representation of the 3D environment (Fig.~\ref{fig:3dgraphs}A).
The resulting graph consists of \num{38392} vertices and \num{639582} edges weighted by Euclidean distance. Based on its spatial extent, the graph diameter is lower-bounded by the number of edges along its longest dimension, which is 77 in this case; the measured diameter is 78. This value is smaller than that observed in comparably sized synthetic 3D grid networks (Fig.~\ref{fig:F2}C), suggesting that the increased number of edges may be compensated by a reduced number of algorithmic iterations.
Indeed, it took for NEURO-MAPP on average \qty{6.1}{\milli\second} to compute an SSSP query on this graph (see Fig.~\ref{fig:3dgraphs}B), compared to \qty{8.0}{\milli\second} for a similar-sized 3D grid network.
A notable distinction from regular 3D grid networks is given by the variability in vertex degree, as vertices adjacent to buildings have fewer neighboring connections. To mitigate potential load imbalances across processing cores of SpiNNaker 2, a vertex-degree-aware partitioning strategy is employed (cf. Section~\ref{sec:met:impl:part}).

In contrast, Dijkstra's algorithm exhibits reduced performance on this data compared to synthetic 3D grid networks, with an average runtime of \qty{6.8}{\milli\second}.
Overall, NEURO-MAPP demonstrates superior efficiency, solving the aerial navigation problem significantly faster, while consuming only \qty{2.4}{\milli\joule} per SSSP query, compared to \qty{25.5}{\milli\joule} required by the CPU-based implementation of Dijkstra's algorithm.

\subsubsection*{Shortest paths in protein interaction networks}

\begin{figure}
    \centering
    \includegraphics[width=\linewidth]{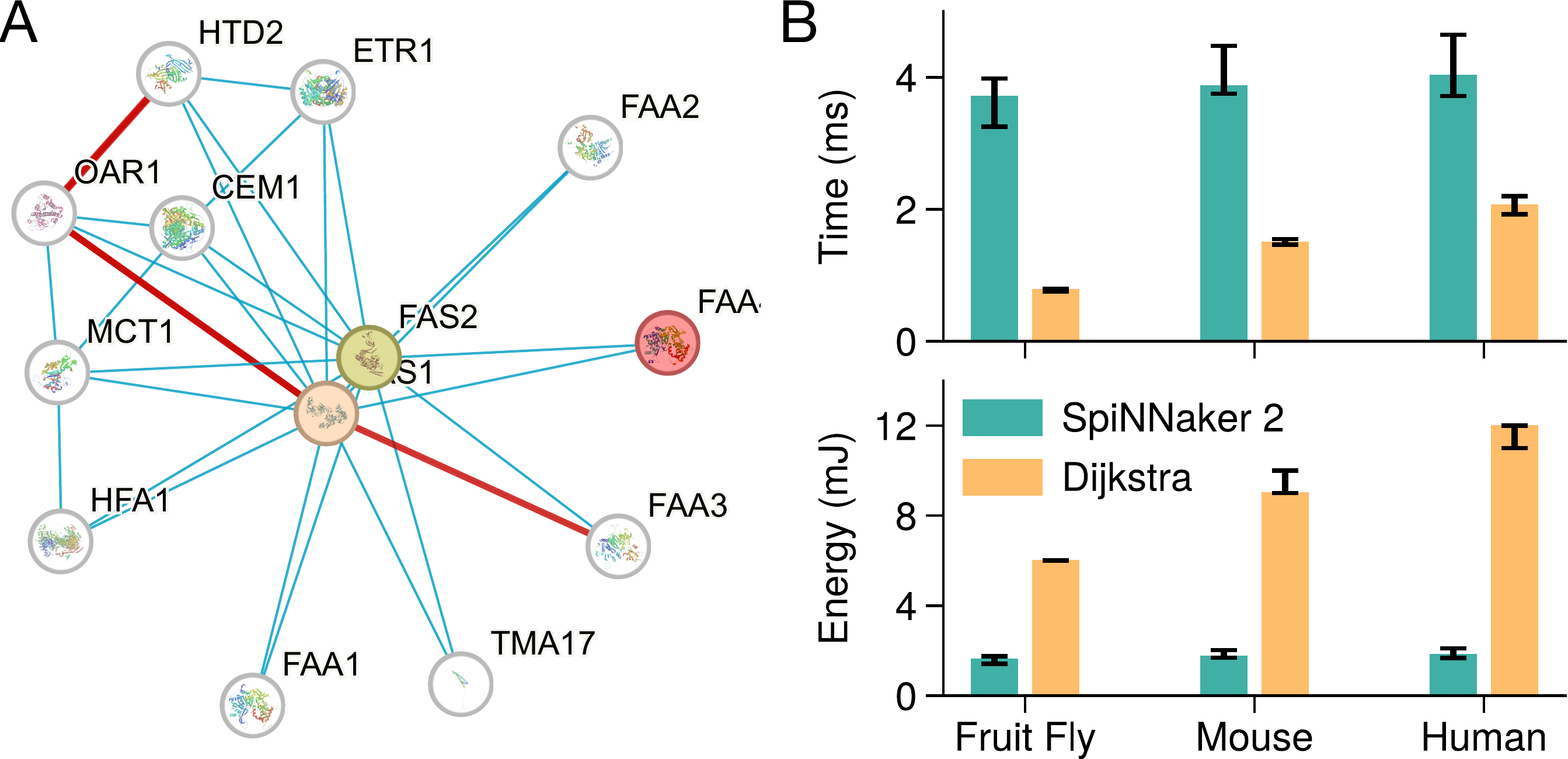}
    \caption{Shortest path search in protein-protein interaction networks. \textbf{A} Example of a small subnetwork, showing a characteristic distribution of vertex degrees in the data, with few hub vertices with high degree and most proteins only having a low number of interaction partners (visualization adapted from \url{https://string-db.org/cgi/network?identifier=4932.YMR246W}, original licensed under CC BY 4.0). \textbf{B} Runtime and energy measurements for both considered algorithmic implementations on protein interaction networks of three different species. 20 trials on randomly selected sources for each network for runtime and energy measurements, plots show median and interquartile range. Note that the network size of the fruit fly is the smallest, while the protein-interaction network of humans is the largest considered (see Table~\ref{tab:protein}).
    }
    \label{fig:protein}
\end{figure}

In biological research, a set of experimentally validated interactions between proteins are commonly represented by protein-protein interaction (PPI) networks (Fig.~\ref{fig:protein}A), with proteins being represented as vertices and their interactions as edges.
We examined such networks for three species, namely fruit fly, mouse, and human, sourced from the STRING database~\cite{szklarczyk_string_2023} and considering only interactions with a confidence score of at least 90\%. 
Given this criterion, the largest available single-species PPI network corresponds to human (``Homo Sapiens'') with \num{11623} proteins.
It is important to note that these networks exhibit highly connected “hub” proteins, some of which interact with hundreds of other proteins. In the human interaction network, for instance, the protein with the highest degree has 372 interactions (see Table~\ref{tab:protein}). 
To efficiently process such networks, we employed the previously mentioned vertex-degree-aware mapping for SpiNNaker 2 (see Methods section~\ref{sec:met:impl:part}).

While the results in Fig.~\ref{fig:protein}B indicate that NEURO-MAPP is slower than Dijkstra's algorithm, we can anticipate from the trend across species (cf. Fig.~\ref{fig:protein}B and Table~\ref{tab:protein}) that for larger graphs with comparable topologies, NEURO-MAPP would outperform the Dijkstra implementation in runtime.
Furthermore, despite longer runtimes, the NEURO-MAPP implementation on SpiNNaker 2 results in a lower energy consumption compared to the CPU-based implementation of Dijkstra's algorithm. 

\begin{table}[!ht]
    \centering
    \caption{Characteristics of the PPI networks considered in this study.}\label{tab:protein}
    \begin{tabular}{lrrr}
    \hline
        \textbf{Species} & \boldmath$|V|$ & \boldmath$|E|$ & \boldmath$\max \deg$ \\ \hline
        \textit{Fruit Fly} & \num{5564} & \num{94340} & 243 \\
        \textit{Mouse} & \num{9693} & \num{145796} & 268 \\ 
        \textit{Human} & \num{11623} & \num{196812} & 372 \\ 
    \end{tabular}
\end{table}

\subsubsection*{Isomap nonlinear dimensionality reduction}

\begin{figure*}
    \centering
    \includegraphics[width=\linewidth]{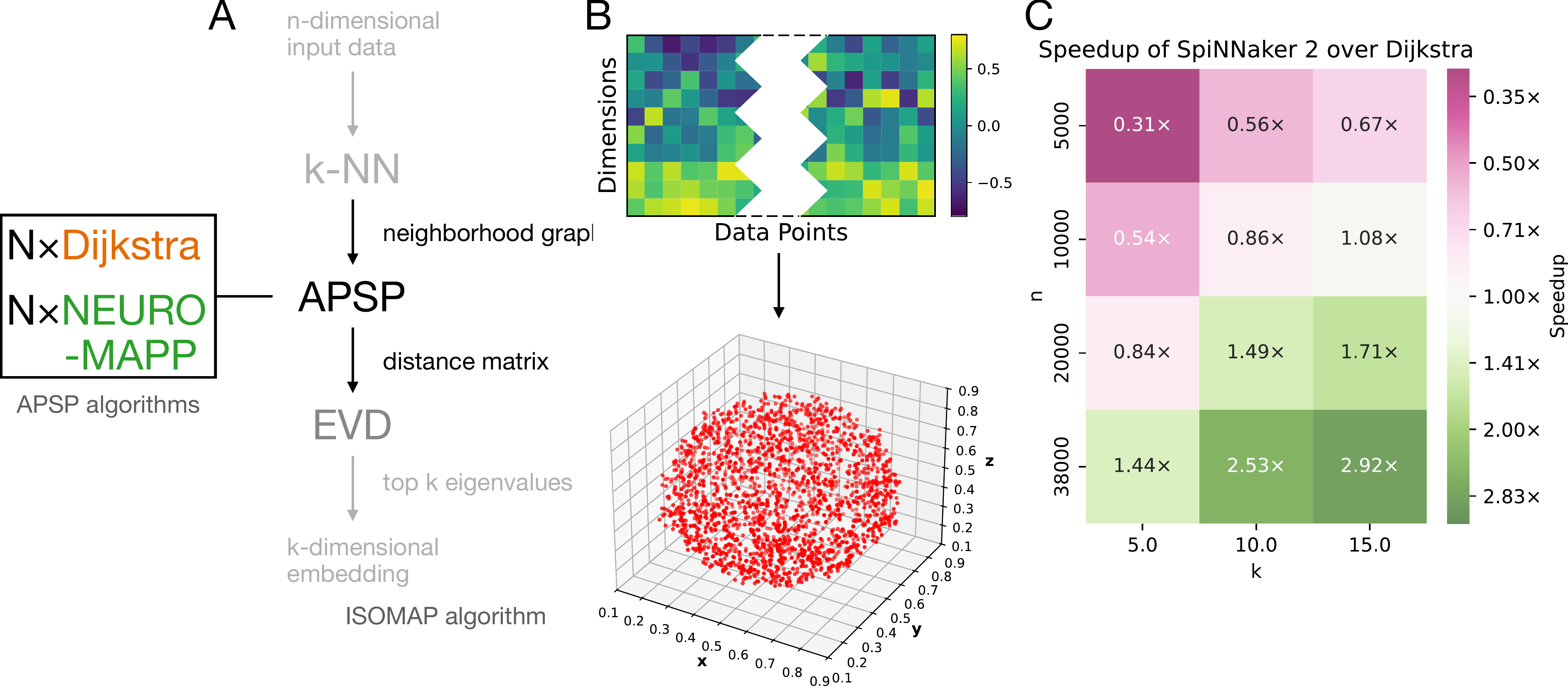}
    \caption{Shortest path finding as part of the Isomap algorithm. \textbf{A} Overview of the different steps involved in the Isomap algorithm and the two considered implementation options for computing the distance matrix of shortest paths. \textbf{B} Exemplary dataset with 10-dimensional input and the three-dimensional embedding decoded by Isomap. \textbf{C} Speedup of NEURO-MAPP relative to Dijkstra's algorithm for different combinations of  number of data points $n$ and considered number of neighbors $k$.}
    \label{fig:isomap}
\end{figure*}

The search for shortest paths is frequently embedded within larger algorithmic frameworks. Using the Isomap method~\cite{tenenbaum_global_2000} as example, we demonstrate how NEURO-MAPP can be integrated into and improve such a framework. Isomap is a mathematical framework for nonlinear dimensionality reduction of large-scale datasets and has been applied in domains including image recognition, speech recognition, or motor planning.

The Isomap framework comprises three principle steps: First, for each data point in the high-dimensional space, the algorithm identifies its nearest neighbors and constructs a weighted graph in which vertices correspond to data points and edges encode pairwise distances between neighboring points.
Second, the algorithm computes the matrix of shortest-path distances between all pairs of vertices in this graph. Third, the centered distance matrix is decomposed, and the $N$ eigenvectors associated with the largest eigenvalues are selected to form the $N$-dimensional embedding of the dataset.

The second step of the Isomap framework contributes substantially to the overall runtime, as it requires the computation of shortest paths between \textit{all} pairs of vertices in a sparse, locally connected graph.
This so-called all-pairs shortest paths (APSP) problem is addressed on CPUs using two approaches. The first employs dedicated APSP algorithms, such as the Floyd-Warshall algorithm.
However, for sparse graphs of the type encountered in Isomap, it is generally more efficient in practice to execute Dijkstra’s algorithm independently from each vertex, thereby computing the distance matrix row by row. NEURO-MAPP, executed on SpiNNaker 2, can be incorporated into the Isomap framework in a similar manner (Fig.~\ref{fig:isomap}A).

To demonstrate the integration of NEURO-MAPP into the Isomap framework and to evaluate its performance relative to Dijkstra's algorithm, we generated a large-scale dataset consisting of points sampled from a 3-dimensional sphere and projected into a 10-dimensional space. The projection was done by generating an orthonormal matrix with added noise. The Isomap framework was subsequently applied to recover the underlying 3-dimensional structure of the dataset (Fig.~\ref{fig:isomap}B).

Within this setup, the second step of Isomap, corresponding to the graph-search computation, was executed on-chip using NEURO-MAPP, whereas the other two steps of the framework were carried out on a CPU.
To assess performance, we measured runtimes for different combinations of the total number of data points $n$ and the number of considered nearest neighbors $k$, which is a key parameter of Isomap. For each combination, we compared the runtime of Dijkstra's algorithm and NEURO-MAPP.
The reported runtimes correspond solely to computational execution and exclude any overhead associated with data transfer or data transformation in either implementation.

Overall, NEURO-MAPP running on SpiNNaker 2 demonstrates superior performance for larger datasets and/or higher values of $k$ (see Fig.~\ref{fig:isomap}C). 
Additional details and further examples are provided in the Supplementary Material (Supplementary Figs.~14 and~15).

\section{Discussion}\label{sec.conclusion}

In this work we have developed and applied the parallel algorithm NEURO-MAPP for SSSP graph search and evaluated its performance on the neuromorphic platform SpiNNaker 2.
We have demonstrated that for many types of graphs, NEURO-MAPP
outperforms the Dijkstra's algorithm, executed on a modern CPU, in terms of runtime and energy consumption. 
While comparing a sequential algorithm for a single core (such as Dijkstra's algorithm) to a parallel algorithm utilizing many cores (e.g., the 152 cores of a SpiNNaker 2 chip) may seem unfair, in the end the important point is how much runtime and energy the different systems require to fulfill a given task. The literature provides a number of such comparisons of runtime and energy use across hardware systems (see, for example, \cite{davies_advancing_2021, blouw2019benchmarking, yan2021, rao2022long, ostrau2022benchmarking, parpart2023implementing, nazeer2024language}). 
Please note that existing parallel SSSP algorithms for CPUs and GPUs are typically less efficient than Dijkstra's algorithm~\cite{davidson_work-efficient_2014}. Therefore, we expect them also to be less efficient than NEURO-MAPP.

Our theoretical estimate indicates that also NEURO-MAPP should be less efficient than Dijkstra's algorithm (cf. section `Energy Consumption' above). Nevertheless, using only a single SpiNNaker 2 chip, NEURO-MAPP outperforms Dijkstra's algorithm on a modern CPU in terms of energy usage on all evaluated graph types, as well as in terms of runtime on a most graph types. Hereby, an important factor is that the cost of the fully utilized system is much lower for SpiNNaker 2 than for the CPU implementation (less than \qty{500}{\milli\watt} compared to \qty{4000}{\milli\watt}).

We further estimated the expected performance on larger graphs that can be targeted with multi-chip SpiNNaker 2 systems, for which the technical conditions were not yet given at the time of our study. 
In the future, our predictions for the performance on larger graphs can seamlessly be tested on multi-chip systems, in particular, on the 48-chip SpiNNaker 2 board~\cite{gonzalez_spinnaker2_2024}, as our implementation can be adjusted for this purpose at relative ease. 

Although the implementation we provide in this work is specific to the SpiNNaker 2 platform~\cite{mayr_spinnaker_2019}, any parallel computing system satisfying the following properties could be used to efficiently implement NEURO-MAPP:

\begin{itemize}
    \item \textit{Sufficient memory local to each processing unit.} This avoids moving graph connectivity and edge cost data through the entire system.
    \item \textit{Ability to run independent programs on each processing unit}. The amount of work to be done at each phase of the algorithm varies between cores. Thus, to optimize runtime and energy efficiency, it is important that cores can work at their own pace and enter a low-energy sleep mode once their work for the current iteration is finished.
    \item \textit{A communication network between the processing units that is able to transmit distance updates efficiently to their target cores.} Ideally this is flexible enough to allow any pair of cores to communicate with each other. However, as we have shown for some graphs it could be sufficient to have communication limited to some local neighborhood.
\end{itemize}

Examples of other hardware platforms satisfying these conditions would be the Graphcore IPUs or other digital, flexibly programmable neuromorphic platforms such as Loihi 2~\cite{jia_dissecting_2019,orchard_efficient_2021}.
Note that prior work has provided first results on the implementation of other SSSP graph search algorithms on those systems~\cite{davies_advancing_2021,gepner_performance_2024}.

While we have presented here an implementation for parallel hardware that allows relatively flexible programming, variants of our algorithm may be used for hardware that is tailored for more specific purposes.
In particular, our algorithm may be adjusted for hardware units dedicatedly designed to execute the dynamics of threshold-gated neurons. This would enable implementations on analog/mixed-signal neuromorphic hardware such as BrainScaleS-2 \cite{pehle2022brainscales}, DYNAP-SE2 \cite{richter2024dynap}, or BrainDrop \cite{neckar2018braindrop}, or on custom application-specific integrated circuits (ASICs), which might allow to run the algorithm in an even more efficient manner for particular settings.
Such an implementation could consist of two main components that perform the computations needed to find the shortest paths. 
Each edge of the graph would be replaced by a circuit of specialized neurons constituting an ``add'' component that adds the cost of the edge to the cost of the path leading up to that edge (Fig. \ref{fig:am_schematic}E) -- in previous work~\cite{von_seeler_adding_2025}, we have already explored possible mechanisms for implementing binary addition using according spiking neural circuits.
Further, each node of the graph would be replaced by a circuit of neurons that constitutes a ``minimize'' component, computing the minimum of the path costs arriving from each of its incoming edges. Such circuits could be implemented based on previous designs using threshold gates~\cite{siu_depth-size_1991,beiu_circuit_1998},\cite[p.303]{kent1992encyclopedia}.
The minimum would then be passed on to the subsequent edges, which add their own cost, and the steps are repeated until all edges have been traversed or a maximum number of iterations is reached ($k$-hops version of the algorithm).

One of the applications we have discussed above, where NEURO-MAPP achieves faster performance than Dijkstra's algorithm, is solving the APSP problem (i.e., computing the shortest paths between all pairs of vertices) as one step of the Isomap framework.
However, transferring the final path distances, which are stored spread over all cores, from the SpiNNaker 2 chip back to the host system represents a bottleneck for the execution time of NEURO-MAPP.
Thus, for an efficient implementation of Isomap utilizing the SpiNNaker 2 system, frequent transfers of data during the runtime of the algorithm should be avoided. One step towards reducing the data transfer frequency would be to coordinate all shortest path queries on the chip and store the resulting distance matrix in the DRAM attached to each chip. This can be implemented with our scheduling system (see Section~\ref{sec:dynamic}).
To further minimize data transfer, other steps of the Isomap framework could potentially be executed on the chip as well. Computation of eigenvectors of the resulting distance matrix could also be optimized using the MAC accelerator for matrix multiplication that is part of each processing element on the SpiNNaker 2 chip.

In addition to considering a single source vertex, NEURO-MAPP supports specifying more than one vertex in the graph as a source. Doing this will cause the algorithm to compute for each vertex the shortest distance from the closest source vertex. By interpreting the sources as points of interest, it also becomes possible to compute the shortest path to the nearest point of interest. To do this, all edges in the directed graph are flipped, changing the interpretation of source vertices to ``targets'', and thus solving the single-destination shortest path (SDSP) problem.
While other SSSP algorithms such as Dijkstra's algorithm allow for the same extension, in our case, using multiple sources accelerates the runtime of the algorithmic implementation -- as the activity starts at multiple positions, the full graph is covered in fewer iterations. Note that as mentioned above, NEURO-MAPP can even efficiently solve the generalized version of the SSSP and SDSP problems, i.e., the APSP problem.

Algorithmically, NEURO-MAPP exhibits similarity to previously developed algorithms for parallel CPU systems. However, in combination with the SpiNNaker 2 hardware, there are essential differences that we will discuss in the following.
In the terminology of~\cite{bertsekas_parallel_1996}, NEURO-MAPP is a ``Parallel Multiple-Queue'' label-correcting algorithm with a few modifications.
The first difference is that we do not require a shared memory between processing units (cores), and instead implement all communication through message passing between pairs of cores. This enables the use of individual queues that are accessed exclusively by their related core, avoiding the need to use locks for accessing shared data structures.
This approach further benefits from the very low communication latency of the SpiNNaker 2 architecture of 12.5 ns per hop \cite{hoppner_spinnaker_2022}. Another difference between NEURO-MAPP and the mentioned Parallel Multiple-Queue algorithms is that we statically assign a subset of vertices to each core.
To keep operations synchronized across cores, we completely fill and empty the queues in rounds, starting at the same time on each unit. 
Furthermore, on the level of single vertices, NEURO-MAPP is similar to the Pregel-based SSSP implementation presented in~\cite{malewicz_pregel_2010}, which is designed for clusters of distributed CPU systems while delegating the details of grouping vertices and handling messages passed between vertices to the Pregel framework. 

Taken together, NEURO-MAPP demonstrates the utility of massively parallel neuromorphic hardware for highly efficient and scalable graph search. Executed on the SpiNNaker 2 chip, NEURO-MAPP reveals strong benefits in energy efficiency and runtime over sequential CPU implementations. Moreover, our results indicate further advantages if NEURO-MAPP is scaled up to larger systems using SpiNNaker 2 and similar platforms.

\section{Methods}\label{sec.methods}

\subsection{Algorithm}\label{subsec.algorithm}

In the following, we provide an abstract description of the algorithm (see Figure~\ref{fig:am_schematic}D,E for examples demonstrating its execution), while the subsequent section elaborates on details of the implementation.

Each vertex in the graph is mapped to an algorithmic unit that holds an internal state, that is, a single value representing the distance estimate to the corresponding vertex from the (closest) source vertex in the graph. For source vertices, the internal state is initially 0, while all other estimates start at $\infty$. In the first iteration, the source vertices send a first distance estimate to their neighbors along the edges of the graph (this estimate equals the cost of the respective edge). In all further iterations, all units whose distance estimate has lowered (i.e., improved) send the new distance estimate to their neighbors, increased by the cost of the connecting edge.
For each such distance update, a unit computes the minimum of its internal state and the incoming distance estimate. If that result is lower than the internal state, the state is updated and the new distance estimate is again sent to the downstream-connected units, increased by the cost of the respective edge.
Assuming exclusively non-negative edge costs, the distance estimate for any vertex $v$ in the graph eventually converges to the true shortest distance between a source $s$ and target $v$ in as many iterations as there are hops in the shortest path between $s$ and $v$. This number of hops is at most $|V|$, as no shortest path will pass through the same vertex twice.

As mentioned above (see section Discussion), previous implementations of SSSP algorithms following this abstract scheme include the class of label-correcting algorithms presented by~\cite{bertsekas_parallel_1996} and an algorithm implemented to demonstrate the capabilities of the \textit{Pregel} graph processing framework~\cite{malewicz_pregel_2010}.
The particular considerations required for our algorithmic implementation on the SpiNNaker 2 chip will be treated in the following section.

\subsection{Implementation on SpiNNaker 2}\label{subsec.impl}

In this section we describe implementational details of our algorithm on SpiNNaker 2. 
After a brief general description, the following subsections cover each point in detail.

Firstly, since the partitioning of vertices across cores has an impact on the performance of our implementation, we developed multiple optimization strategies for graph deployment.
The structure of the graph has to be stored in the SRAM of the SpiNNaker 2 processing elements to provide each core with the information necessary for local computation. The storing of the edges connected to vertices on the core and the storing of incoming distance update information compete for the limited available SRAM.
This constitutes a trade-off between memory and computational efficiency and can impose specific limitations on the graph structure (e.g., restricting maximum vertex degrees). Therefore, we compare two options for the message passing (subsection \ref{methods.impl.mp}): one with slower neighbor lookup but lower memory footprint, and one with fast neighbor lookup at the cost of higher memory usage for each edge.
In general, a table of all edges relevant to one core is generated and stored prior to execution. Depending on the used method, this table contains either all incoming or all outgoing edges of vertices that are assigned to the specific core.

In addition to and depending on the message passing paradigm, we also attempt to optimize the partitioning of vertices across cores directly (subsection \ref{sec:met:impl:part}).

If the duration of each iteration remains fairly constant, then iterations can be run on a fixed interval, which is the default case in the SpiNNaker 2 software stack. However, while the number of algorithmic units is constant for graph search, their update frequency varies significantly over the execution of a single SSSP query. To address this, we implement a dynamic iteration scheduling method (subsection~\ref{sec:dynamic}).

\subsubsection{Message Passing}\label{methods.impl.mp}

SpiNNaker 2 provides a \emph{network-on-chip} (NoC) for communication between cores~\cite{hoppner_spinnaker_2022}. The network is split into two independent parts, the configuration NoC (CNoC) and the data NoC (DNoC). Messages sent over the DNoC contain a 32-bit header specifying the type of packet and information about the target to facilitate the packet routing. Attached to that header is a payload of 32 to 160 bits.
For implementing our algorithm, we make specific use of the DNoC to transmit distance updates between cores. As indicated above, we implemented two different methods for the encoding of those updates and for the tables storing the graph connectivity, which will be presented below.

\textbf{Processing edges at the successor vertex.} 
This method is based on the default implementation of spiking neural networks on SpiNNaker 2, where two hierarchical levels of communication tables are used. On a coarse level, one table stores connections between cores. If any single edge connects a vertex from one core to a vertex on another core, these two cores are considered to be connected.
If a distance update is created at a vertex, a message containing the global index of that vertex and the new distance is sent to all cores that are connected to the one in which the vertex resides.
Then, on the receiving core, the index is looked up in a table containing all incoming edges for the vertices on that core.
The table lookup is implemented with binary search, taking $\mathcal O(\log k)$ time, where $k$ is the total number of vertices in the graph that have an edge to the core.
If there is an entry for the specific predecessor vertex, all distances of successor vertices on that core are compared to the incoming distance increased by the specific weight of that edge.
It is also possible that there is no edge from the given predecessor vertex, as core-to-core connectivity is given if there is any edge between two vertices on the cores. In that case the table lookup fails and the next update is processed.
This method is advantageous for denser graphs, where each possible predecessor has multiple successors on one core. It allows denser storage as only the predecessor has to be stored with a global index, and the successors can be stored with a small address local to the core. Further, one packet sent through the communication network might reach multiple successor vertices, reducing traffic in the network.
On the other hand, for sparse networks or networks with high core-to-core connectivity, the table lookup using a binary search might be too expensive. Especially when there are many vertices with no successor vertex on the receiving core, this will result in a lot of time spent for table searches that find no results.
Further, the many extra spikes to be processed in cases with high core-to-core connectivity cause the spike buffers -- which are responsible for temporarily holding spike messages not yet processed -- to fill up, losing any additional spikes. This can result in obtaining overly long distances, as new, shorter paths are missed, meaning that the correct shortest paths will not be found.

\textbf{Processing edges at the predecessor vertex.}
To reduce processing times, we implemented this second method of distance update processing. Instead of having two levels of connectivity tables, this method only uses one level and performs all relevant processing at the predecessor.
Once a distance update is created, all outgoing edges are looked up in the edge table on the core.
As this table has exactly as many entries as there are vertices on the core, the table lookup is direct and does not require a binary search.
Each table entry contains the list of outgoing edges from that vertex. In turn, each stored edge contains a global address of the successor and the weight of the edge.
For each edge, the weight is added to the updated distance of the predecessor vertex and a message containing the sum and the local address of the successor vertex is sent to the target core.
This method has higher memory requirements per stored edge, as each edge has to store a global vertex address, but replaces the binary search with a constant time lookup.

Larger graphs of some types, like small-world and randomly connected graphs, can not be processed with the default successor-based method due to overflowing spike buffers. Nevertheless, we provide a performance comparison for the two methods for two-dimensional grid graphs in Supplementary Fig.~17. This comparison shows a significant performance advantage for the predecessor-based routing, which grows with the graph size. For grid graphs with \num{\sim 20000} vertices, the total search time is reduced by a factor of $10$.

\subsubsection{Vertex Partitioning}\label{sec:met:impl:part}
To parallelize the implementation, we split the processing of the vertices between the many cores on the SpiNNaker 2 system. We limit the number of vertices processed on one core to 256, placing an upper bound on the memory required to store per-vertex information and allowing addressing vertices within a core using an 8-bit address. With that limit, the maximal graph size that can be processed on a single chip is $152 \times 256 = 38912$ vertices.
The optimal way vertices are assigned to processing elements depends on which of the two variants of message passing discussed above is used.

For the successor-based processing it is essential to reduce the number of cores connected to each other. One core being connected to many other cores will result in an increased number of updates to be sent and processed, most of which will result in table lookup misses.
One method of achieving a sparse core-to-core connection graph is to permute the vertices such that the adjacency matrix has minimal bandwidth, and to assign adjacent chunks of vertices to cores.
While exact minimization of the bandwidth is computationally infeasible for larger graphs (NP-complete~\cite{papadimitriou_np-completeness_1976}), approximation methods produce decent results.
When using successor-based processing, we rearranged all graphs using the \emph{reverse Cuthill-McKee} algorithm~\cite[Chapter 4.4.1]{george_computer_nodate}, and then assigned to cores in groups of 256 nodes. This substantially reduces the required number of core-to-core connections, which in turn reduces total traffic and reduces unnecessary processing of messages (see Supplementary Fig.~8A for an example).

For the predecessor-based processing, optimized vertex partitioning is not required for reducing the number of messages sent, as each distance update is sent to each successor vertex directly instead of all connected cores. However, the partitioning still influences the performance. If we use the same \emph{reverse Cuthill-McKee} ordering as mentioned above, activity in one iteration tends to stay confined to a few active cores, while many other cores are not receiving any updates. Instead, if we randomly assign vertices to cores, activity is spread across all cores more evenly, reducing the total runtime (see Supplementary Fig.~8B).
On the other hand, for graphs with wide distributions of vertex degree, the random assignment to cores may not be ideal. Especially in the case of graphs with few high-degree vertices, protein-protein interactions being one example, there is a chance that multiple of these hubs get assigned to the same core, causing an increase in computational load on that core. For these types of graphs we assigned vertices to cores in a way that minimizes the cumulative degree of all vertices on a core. 
Specifically we performed vertex-degree-aware partitioning using a greedy algorithm to assign vertices to cores. 
Starting with the vertices of highest degree, each vertex is assigned to the core $c_i$ with the current lowest cumulative degree, given by
\begin{equation*}
    \sum_{v \in c_i} \deg^+(v) + \deg^-(v).
\end{equation*}

We want to minimize both in-degree and out-degree, as vertices with high in-degree lead to an increase in incoming distance updates that need to be processed, while vertices with high out-degree require more space in the edge table and cause more time being spent for sending spikes. Further, for the processing of denser graphs, we allow using more cores than strictly necessary according to $\lceil |V| / 256 \rceil$. The effect of this method for a protein interaction graph is shown in Supplementary Fig.~9. Further, we investigated the effect of using all cores for the smaller, sparse graphs we looked at in Fig.~\ref{fig:F2} that can be searched on fewer cores. Results are presented in Supplementary Fig.~10. We observed that while using more cores is slightly less energy-efficient, the difference is minimal as higher power draw is offset by shorter processing time. A visual example of the different mapping strategies is provided in Supplementary Fig.~7.

\subsubsection{Dynamic Iteration Length}\label{sec:dynamic}
When simulating spiking neural networks, SpiNNaker 2 uses a fixed duration for simulating each discrete time step, typically \qty{1}{\ms} for physical ``real-time'' simulations. When a core finishes all of its tasks before the end of that time, it enters a low-power state to wait for the remaining time to pass. If a core takes longer than the allowed time, it starts the next iteration later than the other cores, losing the synchronization, which potentially leads to different, incorrect results.
Choosing the duration of the time step such that all cores can finish the work in every iteration while not waiting too long to reduce the total runtime is essential.
For the SSSP graph search algorithm the amount of work in each iteration is highly variable. For the first and last few iterations only a few updates are processed, while there might be iterations where vertices across the entire graph update their distance (see Supplementary Fig.~18A for an example). With a fixed duration for all iterations that is long enough to prevent incorrect results, any core would spend most of the time waiting for the end of the iteration. This significantly increases the wall-clock time and makes measuring energy consumption by sampling current draw difficult.
While others have implemented dynamic scheduling of iterations on SpiNNaker 2~\cite{langer_octopuscheduler_2025}, no such implementation is available for adoption in our algorithm. 
We thus implemented dynamic scheduling as described in the following.

One of the cores takes the role of the scheduler. This core processes vertices and their distance updates like all other cores, but behaves differently once it is done with this task. When a normal core finishes all graph processing within one iteration, it sends the number of incoming updates it processed in the iteration to the scheduler core, then enters a low-power state waiting for interrupts (a sequence diagram is provided in Supplementary Fig.~18B).
The count is sent using the direct memory access (DMA) feature, writing one 32-bit word to a specific address within the scheduler's SRAM.
The scheduler, configured with the number of total cores used, waits for all cores to send the counts, accumulating all counts. Based on the sum of counts a decision is made how to proceed. If the sum is non-zero, or it is currently the first iteration, all cores should be advanced to the next iteration of the algorithm. This is done by sending a feed-through interrupt, a signal that wakes all cores from the low-power mode, calling the callback function defined for the specific type of interrupt. In this case, this function starts the execution of the next iteration.
Otherwise, if the sum is zero, the algorithm has converged. Now, the chip can terminate execution.
Alternatively, a new run of the algorithm can be started. For power measurement purposes it can be useful to repeat the exact same run many times to get a consistent load on the chip. In that case, the scheduler sends a different feed-through interrupt to all cores, causing them to reset the distances of all vertices to their initial values.

We note that it would also be possible to change the source node between runs, allowing to efficiently compute all-pairs shortest paths (APSP) without the need for repeated initialization from a host CPU. However, a full implementation of APSP would also require storing the results in the DRAM to build the full distance matrix, which is not yet implemented.

\subsection{Energy and Runtime Measurement}\label{sec:met:erm}

To evaluate our implementation, we compared the results to an implementation of Dijkstra's algorithm. For this comparison we need to be able to measure the runtime and energy consumption of both implementations. For the SpiNNaker 2 system, timings are measured from within our implementation on-chip, while energy is measured through external sensors.

Timings are measured with fixed duration time steps (see Section~\ref{sec:dynamic}). To avoid including time when all cores are waiting for the next iteration, we record the time taken for each core in each iteration. After the algorithm has been run, these recordings are read from the chip. For each time step, we take the maximum of the recorded times and consider this to be the duration of that iteration. The total time we report is the sum over these iteration durations for all iterations until convergence.

The SpiNNaker 2 chip is equipped with sensors for measuring current draw on five different power lines. The sensors can be accessed through the STM Python library~\cite{spinnaker2_stm}, reading a measurement sample every \qty{100}{\milli\second}. The current measurements are converted to power draw values by multiplying with the voltage of the respective power line. We repeatedly run the algorithm, until \qty{5}{\second} have passed, collecting \num{\sim50} samples for the current draw during execution (section~\ref{sec:dynamic} describes how this is implemented). Manually inspecting the samples clearly highlights the five seconds during which the graph search runs by an increased power draw on the two power lines for the CPU cores. The current draw on the other power lines remains constant. To obtain a single number for the power draw of the full chip during execution of the algorithm, the maximum power draw is computed independently for each power line, and the sum over all five maxima is reported (see Supplementary Fig.~19).

The implementation we use for Dijkstra's algorithm for comparison is a \CC\ implementation from the Boost Graph Library, specifically with \textit{dijkstra_shortest_paths} using the adjacency list representation for the graphs. This implementation was executed on a M1 Pro CPU. Time is measured from within \CC, directly before and after the call to this function, excluding any time taken for the construction of the graph data structures. 

The energy consumption of the full CPU during execution of Dijkstra's algorithm is measured using a small library~\cite{zeus_applesilicon} to query energy counters directly before and after execution and compute the difference. This will include some background activity from other processes, however, the power draw of such background activity is typically two orders of magnitude lower than that of running Dijkstra's algorithm. Outliers with high background activity were discarded in Fig.~\ref{fig:F2} by considering the median energy consumption.

For each graph parameter setting, we performed multiple runs using both algorithms and recorded their runtimes.
Graph metadata is recorded with the timing and energy measurements to allow reconstruction of equal or comparable test cases on the other platform.
When running on SpiNNaker 2, the resulting distances are read from the chip, and verified for correctness using a reference SSSP implementation. For further analysis, the unweighted diameters of these graphs are computed with \CC\ and added to the database. These measurements are shown in Supplementary Figs.~1-6.

\subsection{Graph Data}\label{sec.methods.graphs}

This section describes the retrieval or generation of the graphs used for evaluating the algorithms in the results section. Further, characteristic properties such as number of vertices, edges, degree, and graph diameter are given.

\subsubsection{Synthetic Graphs} 

Synthetic graphs were used in the evaluation of our NEURO-MAPP implementation to have fine-grained control over graph characteristics. Multiple types of graphs were generated.
In the case of grid graphs, small-world graphs, and locally connected ring graphs, where the generated graphs are initially unweighted and undirected, we convert them to directed and weighted (bidirectional but asymmetric) graphs, replacing every undirected edge by two directed edges and sampling weights for each edge independently from a uniform distribution over $[0, 10^4]$ (cf. \cite{luboeinski2023stabilization}).

\textbf{Grid graphs} have their vertices on the Cartesian grid $\{1, 2, \ldots, n\}^d$, with $n$ being the side length of the grid and $d$ the number of dimensions. Two vertices are connected by an edge if they differ in exactly one component, by one unit ($(v, w) \in E \Leftrightarrow \| v-w \|_2 = 1$). Interior nodes in these grid graphs have $\deg^+=\deg^-=2d$ neighbors, one for every direction. The total number of nodes in these graphs is $n^d$, with slightly less than $n^d \times 2d$ edges, as the aforementioned interior vertices have exactly $2d$ edges, while vertices on the boundary have fewer.
The longest unweighted path within the graph, or the diameter of the graph, is between nodes $(1, \ldots, 1)$ and $(n, \ldots, n)$ and has a length of $d \times (n-1)$ hops. If we consider the total number of vertices in the graph $|V|$, the diameter is $d \times (|V|^{1/d} - 1)$.

\textbf{Small-world networks} are generated following the Watts-Strogatz model for undirected, unweighted graphs~\cite{watts_collective_1998} using the generator from the \textit{networkx} library.
The Watts-Strogatz model has parameters $n$, $k$, and $p$. Initially, $n$ nodes are connected to their nearest $k$ neighbors, resulting in a ring structure. Then each edge is rewired with probability $p$, replacing one of the nodes in the pair with a random node within the graph, while avoiding duplicate and self-edges.
The in-degree of each node is initially controlled by the parameter $k$, the random rewiring maintains this as an average. As the rewired ``global'' connections serve to skip large parts of the graph, small-world graphs typically have a comparably low diameter.
Typical values for the rewiring probability $p$ are between 0.01 and 0.1. If not mentioned otherwise, we use $p=0.1$.
From the observations that are shown in Supplementary Figs.~1-3, the graph diameter decreases with increases in $k$ or in $p$. And we observed slower growth with an increase in $n$ compared to the grid graphs, leading us to speculate the diameter grows with $\mathcal O(\log n)$, similar to Erd\H{o}s–R\'{e}nyi graphs. This assumption is used to produce the fits in Supplementary Fig.~13, which provide a good model for the measured runtimes~\cite{newman_renormalization_1999}.

\textbf{Locally connected ring graphs} are generated in the same way as small-world networks, but with $p$ set to 0. Without the edge rewiring, every vertex has exactly $k$ neighbors, and the diameter of the graph is $n/k$, with the longest path being between two opposite vertices, following the circular route. With $k>2$, every hop can skip $k/2-1$ vertices, reducing the diameter slightly. See Supplementary Fig.~4 for experimentally measured diameters matching these assumptions.

\textbf{Random graphs} consist of $n$ vertices, each having a fixed out-degree $k$. The edge targets are randomly selected among the other vertices of the graph, without generating duplicate or self-edges. The fixed out-degree results in the number of edges $|E| = k|V|$.
The diameter of these graphs is generally very low, in the range between 1 and 10 for the graphs we considered (see Supplementary Fig.~6). For interpolation of runtime we again assumed $\mathcal O(\log n)$ diameter growth, which allowed us to match the observed runtime in Supplementary Fig.~12.

\subsubsection{Real-World Graphs}\label{sec.methods.graphs.real}

For evaluating the implementation on real-world graphs, we considered 2D road networks, 3D aerial navigation, protein interaction networks, as well as non-linear dimensionality reduction with Isomap.

\textbf{Road networks} were fetched from OpenStreetMap. A list of districts (Landkreis and city districts) in Germany was obtained through the Overpass API. For each district, the network of roads was obtained with the use of the \textit{osmnx} library~\cite{boeing_modeling_nodate}.
Intersections in the networks were simplified to single vertices in the graph.
As these districts are not large enough to fill a SpiNNaker 2 chip, we generated larger map graphs by sampling coordinates across Germany, and created maps with increasing radii around these coordinates.
These graphs are quite sparse, typically having between 2.4 and 2.8 directed edges per vertex.
The diameters, computed using repeated BFS searches through unweighted copies of the graph, were in a similar range as two-dimensional grid graphs, scaling approximately with $\sqrt{|V|}$ (see Supplementary Fig.~20).
Differing from the synthetic graphs, however, the edge weights do not follow a uniform distribution. Instead, they are determined by the length of the corresponding road segment, rounding to the nearest whole centimeter.
The distribution of these weights follows the distribution of distances between intersections within the selected map area.

\textbf{3D environment}: Using the Blosm plugin~\cite{blosm_blender}, building data for an area of $\qty{1347}{\meter} \times \qty{1930}{\meter}$ in Manhattan was downloaded from OpenStreetMap into Blender, and converted to a 3D mesh. Additional meshes for roads and environment were manually removed to just keep the building data. The meshes were then closed by adding a ground plane, and exported for further processing. Using the trimesh library in Python, a regular grid of points with \qty{25}{\meter} spacing was placed within the horizontal bounding box, with a vertical range of 0 to \qty{250}{\meter}. Points that ended up inside of buildings were removed. An edge was added between neighboring points -- including diagonals -- if the direct path between the points was not obstructed by a building. Euclidean distances along these connections were used as edge costs. The resulting graph containing all possible paths has \num{38392} vertices and \num{639582} edges with a maximum vertex degree of 26 and a diameter of 78.

For the \textbf{protein interaction networks}, we downloaded the data from the STRING database for the organisms \emph{Drosophila melanogaster}, \emph{Mus musculus}, and \emph{Homo sapiens}. Only interactions with \emph{Highest confidence} ($\ge 90\%$) were included. The inverse confidence of an interaction was used as the edge weight. All proteins not connected to the largest connected graph component by any interaction were discarded.

For \textbf{Isomap}, we evaluated two datasets (see Fig.~\ref{fig:isomap} and Supplementary Figs.~14,15). The first dataset is based on the generation of a 3-sphere, which is then embedded into 10 dimensions using a randomly generated, orthonormal embedding matrix, and finally the points are offset by normally distributed noise.
Secondly, the Swiss Roll dataset~\cite{Marsland2014MachineLA} is created by sampling random points on a ``rolled up plane''. Isomap is expected to return an unrolled representation of that plane, with neighbors in the plane maintaining their relative position. In both cases, graphs are constructed from the point clouds by connecting each point to its $k$ nearest neighbors, with the Euclidean distance being used as edge weights.

\section*{Acknowledgments}
The authors thank SpiNNcloud Systems and the Gesellschaft für wissenschaftliche Datenverarbeitung mbH Göttingen (GWDG) for providing access to SpiNNaker 2 resources.
This work was funded by a resources grant from the Campus Institute Data Science (CIDAS), Göttingen, Germany, and by German Federal Ministry of Education and Research (BMBF) grant number 01IS22093A-E.
\section*{Author contributions}

Conceptualization: OS, JL, AL, CT; Data curation: OS; Formal analysis: OS, JL, AL; Funding acquisition: JL, AL, CT; Investigation: OS, JL, AL; Methodology: OS, EO, CM, TK, JL, AL; Project administration: JL, AL, CT; Software: OS; Resources: CT; Supervision: JL, AL, CT; Validation: OS; Visualization: OS, JL, AL; Writing -- original draft: OS, JL, AL; Writing -- review \& editing: all authors.

\section*{Competing interests}

ABL is a co-founder and shareholder of Circulant GmbH. The remaining authors declare no competing interests.

\bibliographystyle{IEEEtran}
\bibliography{references.bib}

\vspace{12pt}

\end{document}